\documentclass{article} 
\usepackage[preprint]{colm2026_conference}

\usepackage{microtype}
\usepackage{hyperref}
\usepackage{url}
\usepackage{booktabs}
\usepackage{amsmath}
\usepackage{multirow,multicol}
\usepackage{adjustbox}
\usepackage[table]{xcolor}
\usepackage[most]{tcolorbox}
\usepackage{wrapfig}

\newtcolorbox{simplebox}{
  colback=blue!5!white,   
  colframe=blue!75!black, 
  coltitle=black,
  fonttitle=\bfseries
}

\usepackage[most]{tcolorbox}

\newtcolorbox{observation}[1]{
  enhanced,
  before skip=4pt,
  after skip=2pt,
  colback=white,
  colframe=blue!50!black,
  boxrule=0.8pt,
  sharp corners,
  fonttitle=\bfseries,
  coltitle=blue!50!black,
  attach boxed title to top left={
    xshift=12pt,
    yshift=-2.5mm,      
    yshifttext=-1.5mm 
  },
  boxed title style={
    colback=white,    
    colframe=white,   
    outer arc=0pt,
    arc=0pt,
    bottom=0pt,       
    top=0pt,          
  },
  title={Observation #1}
}

\usepackage{amsmath,amsfonts,bm}

\def\eqref#1{equation~\ref{#1}}

\def\1{\bm{1}}
\newcommand{\train}{\mathcal{D}}

\def\rvx{{\mathbf{x}}}
\def\rvy{{\mathbf{y}}}

\DeclareMathAlphabet{\mathsfit}{\encodingdefault}{\sfdefault}{m}{sl}
\SetMathAlphabet{\mathsfit}{bold}{\encodingdefault}{\sfdefault}{bx}{n}

\newcommand{\E}{\mathbb{E}}

\newcommand{\KL}{D_{\mathrm{KL}}}

\usepackage{xurl}
\usepackage[capitalize,noabbrev,nameinlink]{cleveref}
\makeatletter
\crefname{section}{\S\@gobble}{\S\@gobble}
\crefname{subsection}{\S\@gobble}{\S\@gobble}
\crefname{proposition}{Prop.}{Props.}
\crefname{figure}{Fig.}{Figs.}
\crefname{equation}{Eq.}{Eqs.}
\crefname{table}{Table}{Tables}
\makeatother

\usepackage[table]{xcolor}
\definecolor{lightestblue}{RGB}{240, 248, 255} 
\newcommand{\highlight}{\cellcolor{white}}

\usepackage{lineno}

\definecolor{darkblue}{rgb}{0, 0, 0.5}
\hypersetup{colorlinks=true, citecolor=darkblue, linkcolor=darkblue, urlcolor=darkblue}

\usepackage{tcolorbox}
\tcbuselibrary{skins} 

\newtcolorbox{titledbox}[1]{
    colback=blue!3!white,      
    colframe=blue!40!black,    
    fonttitle=\bfseries,       
    title=#1,                  
    boxrule=0.5pt,
    arc=3pt
}

\title{Beyond Distribution Sharpening: The Importance of Task Rewards}

\author{
  Sarthak Mittal\thanks{Correspondence to \texttt{sarthmit@gmail.com} and \texttt{guillaume.lajoie@mila.quebec}}\textsuperscript{~~~1,2}, 
  Leo Gagnon\textsuperscript{1,2}, 
  and Guillaume Lajoie\textsuperscript{1,2} 
  \\[0.5ex]
  \small \textsuperscript{1}Mila \quad \textsuperscript{2}Université de Montréal\\
}
\begin{document}

\ifcolmsubmission
\linenumbers
\fi

\maketitle

\begin{abstract}
Frontier models have demonstrated exceptional capabilities following the integration of task-reward-based reinforcement learning (RL) into their training pipelines, enabling systems to evolve from pure reasoning models into sophisticated agents. However, debate persists regarding whether RL genuinely instills new skills within a base model or merely sharpens its existing distribution to elicit latent capabilities. To address this dichotomy, we present an explicit comparison between distribution sharpening and task-reward-based learning, utilizing RL as a tool to implement both paradigms. Our analysis reveals the inherent limitations of distribution sharpening, demonstrating from first principles how and why the optima can be unfavorable and the approach fundamentally unstable. Furthermore, our experiments using \texttt{Llama-3.2-3B-Instruct}, \texttt{Qwen2.5-3B-Instruct} and \texttt{Qwen3-4B-Instruct-2507} on math datasets confirm that sharpening yields limited gains, whereas incorporating task-based reward signal can greatly help achieve robust performance improvements and stable learning.
\end{abstract}
\section{Introduction}
\looseness=-1
The rapid advancement of frontier models has been driven by a shift from next-token prediction to goal-oriented post-training. In particular, task-reward-based Reinforcement Learning (RL) has become a central component of modern large language model (LLM) pipelines \citep{ouyang2022training, schulman2017ppo,shao2024deepseekmath,team2025every,yu2025dapo,lambert2025reinforcement}, enabling strong performance on tasks requiring multi-step reasoning, tool use, and planning \citep{comanici2025gemini,singh2025openai,team2026kimi,yang2025qwen3,jaech2024openai,guo2025deepseek}. Implicit in this paradigm is the hypothesis that RL fine-tuning improves models by enabling the acquisition of new capabilities through optimization with respect to task-dependent rewards. However, despite these empirical successes, the mechanisms underlying these improvements remain poorly understood.

\textbf{Distribution Sharpening Hypothesis}

\looseness=-1
A recent line of work has proposed that many of the gains attributed to RL fine-tuning can be explained by \emph{distribution sharpening} \citep{huang2024self,karan2025reasoning}. Intuitively, this view holds that post-training primarily makes the model more confident in its existing preferences -- it reduces uncertainty and concentrates probability mass on outputs that the model already considers plausible, rather than introducing fundamentally new behaviors. Formally, this is achieved by reweighting the base model's distribution to be more concentrated around its modes, with extreme cases like beam search putting all the mass on a single mode.

\looseness=-1
This hypothesis is supported by recent results showing that inference-time (i.e. training-free) procedures which concentrate probability mass on high-likelihood trajectories can recover strong performance on reasoning tasks \citep{karan2025reasoning, ji2026scalablepowersamplingunlocking}. Similarly, another line of work \citep{shao2025spurious} argues that task-reward is not important and models can improve solely from spurious signals. Furthermore, analyses of RL-based fine-tuning have identified biases toward reinforcing already likely responses and difficulties in discovering or maintaining less probable but valid solutions \citep{fan2026sharpeningcollapsesamplingbias, he2025rewardingunlikelyliftinggrpo}, reinforcing the view that standard post-training methods may primarily act by amplifying existing preferences.
\looseness=-1
At present, it remains unclear to what extent the benefits of RL fine-tuning can be explained by such sharpening effects, or whether optimizing a task-dependent reward provides additional advantages. In particular, these mechanisms have not been compared in a controlled setting where they can be isolated and evaluated on common grounds.

\looseness=-1
Resolving this question has important implications for the design and scaling of post-training methods. If the gains of RL fine-tuning arise primarily from distribution sharpening, then improvements may be achieved through better inference or confidence calibration, and would ultimately plateau based on the quality of the pre-trained distribution. Conversely, if task-reward optimization provides benefits beyond sharpening, then the design of reward signals remains a central component of capability scaling.

\textbf{Approach}

\looseness=-1
To investigate this question, we leverage the standard KL-regularized RL framework used for LLM fine-tuning \citep{ouyang2022training}, which combines a reward maximization objective with a KL divergence term against a target distribution. By varying the contribution of each term, we can express a spectrum of objectives within the exact same training procedure: pure task-reward optimization, distribution sharpening alone, or a combination of both. This ensures that any observed differences arise from the signal being optimized and the underlying optima of the procedure rather than from changes in training paradigms or inference procedures. Using mathematical reasoning tasks of varying difficulty, we compare each of the methods above with each other and against inference-time distribution sharpening baselines.

\begin{table}[t]
\centering
\footnotesize
\setlength{\tabcolsep}{4.5pt}
\begin{tabular}{l|ccccc}
\toprule
\textbf{Method} & \textbf{Name} & \textbf{Reward} & \textbf{KL Target} & \textbf{KL Coeff.} & \textbf{Optimal Policy} \\
\midrule
\multirow{2}{*}{Task-Reward} 
& Optimization 
& $r_{\text{task}}$ 
& -- 
& -- 
& $\arg\max r_{\text{task}}$ \\

& Tilted Sampling 
& $r_{\text{task}}$ 
& $\pi_{\theta_{\text{base}}}^\alpha$ 
& $\beta$
& $\propto \pi_{\theta_{\text{base}}}^\alpha \exp(\beta^{-1}r_{\text{task}})$ \\
\midrule

\multirow{2}{*}{Dist. Sharpening} 
& Optimization 
& $\log \pi_{\theta_{\text{ref}}}$ 
& -- 
& --
& $\arg\max \pi_{\theta_{\text{base}}}$ \\

& Tempered Sampling 
& -- 
& $\pi_{\theta_{\text{base}}}^\alpha$ 
& 1
& $\propto \pi_{\theta_{\text{base}}}^\alpha$ \\
\bottomrule
\end{tabular}
\caption{\textbf{Four configurations considered in this work.} All methods share the same RL training framework and differ only in the reward signal and the presence of KL regularization.}
\label{tab:methods}
\vspace{-4mm}
\end{table}

By isolating these objectives within a unified framework, this work provides a clearer understanding of the mechanisms underlying post-training gains and the role of task-reward-based optimization beyond distribution sharpening.

\textbf{Key Contributions}
\begin{itemize}
\item \textit{A Principled Comparison Framework}: We introduce an RL-based training pipeline that allows for the isolation and controlled comparison of distribution sharpening and reward-based optimization.
\item \textit{Beyond Sharpening on Difficult Tasks}: We show that reward maximization consistently improves performance beyond what can be achieved by sharpening alone, with gains that are particularly pronounced on more difficult tasks where the base model performs poorly. This suggests that the benefits of RL fine-tuning cannot be fully explained by confidence amplification, and highlights the role of task reward in shaping model behavior in regimes where sharpening is insufficient.
\end{itemize}
\section{Background}
In this section, we first explain the fundamentals behind KL-regularized RL based fine-tuning of LLMs. We then discuss the mathematical objectives and optima underlying the two paradigms of interest: task-reward fine-tuning and distribution sharpening. Later, in \Cref{sec:method}, we explain how both can be optimized through the aforementioned RL framework.

\subsection{KL-Regularized RL Fine-Tuning}
\looseness=-1
Let $\pi_\theta$ denote the LLM policy with parameters $\theta$, and $\pi_{\theta_{\text{ref}}}$ denote the \textit{reference} model, often but not necessarily set as the base model $\pi_{\theta_{\text{base}}}$ before fine-tuning. Then, the RL objective can be cleanly described as the following optimization procedure
\begin{align}
    \label{eq:rl-optimization}
    \arg\max_\theta \E_{\rvx \sim \train}\left[\E_{\hat{\rvy} \sim \pi_\theta(\cdot | \rvx)}\left[r(\hat{\rvy}, \rvx)\right] - \beta \KL\left(\pi_\theta(\cdot | \rvx) || \pi_{\theta_{\text{ref}}}(\cdot | \rvx)\right)\right]
\end{align}
where $r(\cdot, \rvx)$ is a scalar reward function which generally depends on the problem $\rvx$, $\train$ denotes the data distribution, and $\beta \geq 0$ is a hyperparameter controlling the degree to which the optima should diverge from the reference model under the Kullback-Leibler Divergence measure. In this setting, $\rvx$ refers to the prompt and $\hat{\rvy}$ to the generations from the LLM $\pi_\theta$. Optimizing the above objective has the following global optima
\begin{align}
    \pi_{\theta^*}(\cdot | \rvx) = \pi_{\theta_{\text{ref}}}(\cdot | \rvx) \text{exp}\left(\beta^{-1} r(\cdot, \rvx)\right)
\end{align}
where $\beta$ dictates the influence of reward on the optimal policy and is used to prevent the model from drifting too much from a potentially generalist reference model. This objective itself has been of considerable interest where a number of works demonstrate improved performance without any KL regularization \citep{chen2025minimax,rastogi2025magistral,khatri2025art,shah2025comedy}, i.e. $\beta = 0$ which implies solely incentivizing maximization of reward.

\looseness=-1
The optimization for \Cref{eq:rl-optimization} can be simply done through gradient ascent on parameters $\theta$, which leads to the REINFORCE estimator \citep{williams1992simple}. In practice, this objective is generally high variance and thus practitioners employ a control variate, commonly called a baseline, which keeps the process unbiased but reduces variance \citep{shao2024deepseekmath,yu2025dapo,kool2019buy,mnih2016variational}. In this work, we limit our scope to the leave-one-out control variate which has been shown to lead to unbiased gradients and stable performance \citep{kool2019buy, tang2025pitfallskldivergencegradient}. For each prompt $\rvx$, this estimator is obtained by unrolling multiple responses $\hat{\rvy}_1, \ldots, \hat{\rvy}_k$ (commonly called group) and considering an empirical average of the reward of $\hat{\rvy}_{-i}$\footnote{$\hat{\rvy}_{-i}$ denotes the set of all trajectories after removing $\hat{\rvy}_i$.} trajectories as the baseline for response $\hat{\rvy}_i$.

We next discuss the objective of both paradigms of interest -- task-reward-based learning and distribution sharpening -- and consequently how they can be optimized with the framework described in this section \Cref{eq:rl-optimization}.
\subsection{Task-Reward Fine-Tuning}
\looseness=-1
This approach remains the predominant method for enhancing LLM capabilities through environment feedback. Formally, it corresponds to defining the reward function $r(\cdot, \rvx)$ in \Cref{eq:rl-optimization} as an evaluation of output quality for a given task. These tasks generally fall into two distinct categories. A task is considered \emph{verifiable} when its correctness can be objectively measured against a definitive ground truth, such as the accuracy of a mathematical solution or executable code. For example, in the case of math problems one can consider $r(\hat{\rvy}, \rvx) := r_{\text{task}}\left(\hat{\rvy}, \rvy(\rvx)\right)$, where $r_{\text{task}}$ checks if the final answer in the generated response $\hat{\rvy}$ matches the final answer for that particular problem $\rvy(\rvx)$. Conversely, a task is \emph{non-verifiable} when the evaluation criteria are subjective -- such as aesthetics or creativity -- requiring rewards to be derived from learned preference models, human feedback, or LLM annotations.
 
\subsection{Distribution Sharpening}\label{sec:dist_sharp}
\looseness=-1
An alternative approach to improving generation quality focuses on eliciting optimal responses from a model without relying on external feedback. Formally, the objective is to extract responses that are more likely under the base model $\pi_{\theta_{\text{base}}}$ than those produced by standard ancestral sampling, $\pi_{\theta_{\text{base}}}(\rvy_{t}\mid \rvy_{<t}, \rvx)$. This is typically achieved either by optimizing to obtain the most likely response, $\arg\max_\rvy \log \pi_{\theta_{\text{base}}}(\rvy | \rvx)$, which is done in beam search \citep{chen2018stableeffectivelearningstrategy} or by sampling from a tempered distribution $\pi_{\theta_{\text{base}}}^\alpha$ (where the temperature is $\tau := 1/\alpha$). Crucially, naïvely applying a lower temperature during ancestral sampling does not equate to tempered sampling over the entire response sequence (\citealp{karan2025reasoning}, \textit{Proposition 1}), necessitating more sophisticated methods. For e.g. power sampling \citep{karan2025reasoning} is an inference-time method that leverages a block-wise MCMC-based sampling to simulate tempered sampling.

\looseness=-1
Beyond inference-time methods, several training-based solutions have been explored. For instance, \citet{hu2023amortizing} proposed learning a policy $\pi_\theta$ that fits a tempered version of the base distribution $\pi_{\theta_{\text{base}}}^\alpha$ using a GFlowNet objective \citep{bengio2021flow, malkin2022trajectory}, yielding improved generation. Similarly, \citet{huang2024self} demonstrated that preference-based RL fine-tuning -- using the model's own likelihood as the preference signal -- can lead to significant performance gains.

\looseness=-1
Coupled with recent findings highlighting the distribution-sharpening tendencies of RL fine-tuning \citep{fan2026sharpeningcollapsesamplingbias, he2025rewardingunlikelyliftinggrpo}, the success of these self-improvement techniques implicitly challenges the unique benefits of external task-reward based RL fine-tuning. This naturally motivates a principled, comparative analysis between the two paradigms.

In the following section, we demonstrate how the task-reward optimization framework can also be used to simulate distribution sharpening. This unified setup enables a controlled comparison between distribution sharpening and task-reward-based fine-tuning, allowing us to isolate and evaluate the true impact of the task reward.
\section{Method}
\label{sec:method}
We can now instantiate the KL-regularized RL objective of \Cref{eq:rl-optimization} with different choices of reward $r$, regularization $\beta$ and reference target $\pi_{\theta_{\text{ref}}}$ for the KL divergence, yielding four regimes corresponding to the combinations of task-reward and distribution sharpening discussed above. We give each a name which we will reuse in the remainder of the paper.

\textbf{Task-Reward RL:} $r = r_{\text{task}}, \ \beta = 0$ corresponds to pure reward maximization.
\begin{align}
    \label{eq:reward-max}
    \pi_{\theta^*} = \arg\max r_{\text{task}}
\end{align}
This is one of the common paradigms for RL fine-tuning on verifiable tasks, and has been heavily explored and leveraged for large-scale post-training of LLMs \citep{chen2025minimax,rastogi2025magistral,khatri2025art}. It can potentially suffer from mode-collapse as it incentivizes the policy to put all its mass on the maximum reward response.

\textbf{Tilted Sampling:} $r = r_{\text{task}}, \ \beta > 0$, with a tempered reference $\pi_{\theta_{\text{ref}}} = \pi_{\theta_{\text{base}}}^\alpha$. 
\begin{align}
    \label{eq:tilted}
    \pi_{\theta^*} \propto \pi_{\theta_{\text{base}}}^\alpha \exp(\beta^{-1} r_{\text{task}})
\end{align}
This corresponds to a tilted distribution. We use a tempered reference model to enable a direct comparison with distribution sharpening, isolating the effect of combining task reward with sharpening. Here, $\alpha =1$ corresponds to normal KL-regularized RL, which has been of most interest in non-verifiable tasks which can suffer from reward hacking \citep{ouyang2022training} and has also been heavily utilized in large-scale post-training recipes \citep{guo2025deepseek,research2026composer,team2025kimi}.

\textbf{Distribution Sharpening:} $r = \log \pi_{\theta_{\text{base}}}, \ \beta = 0$, corresponds to learning a policy that generates responses which maximize the likelihood under the base model, like beam search.
\begin{align}
    \label{eq:dist-sharpen}
    \pi_{\theta^*} = \arg\max \pi_{\theta_{\text{base}}}
\end{align}
\looseness=-1
The optima of this procedure yields a policy that generates sequences with maximal likelihood under the base model, serving as the foundation for distribution sharpening. While beam search shares this objective, its optimization capacity is constrained by its nature as an inference strategy. In contrast, leveraging RL to simulate this process enables more robust optimization by iteratively updating the policy weights to achieve superior convergence.

\looseness=-1
\textbf{Tempered Sampling:} $r = 0, \ \beta = 1$, with a tempered reference $\pi_{\theta_{\text{ref}}} = \pi_{\theta_{\text{base}}}^\alpha$ leads to learning a policy which can sample from the tempered distribution.
\begin{align}
    \label{eq:tempered}
    \pi_{\theta^*} \propto \pi_{\theta_{\text{base}}}^\alpha
\end{align}
This corresponds to learning a policy that bridges between being able to sample multiple responses and providing high likelihood generations under the base model. Such sampling is explored in \citet{karan2025reasoning} from an inference-time perspective, as well as in \citet{hu2023amortizing} to amortize the sampling through a similar fine-tuning process.

We refer to \Cref{tab:methods} for an overview of the different training paradigms considered here.
\section{Analysis}
In this section, we detail the models, datasets, and training protocols used in our study. Our experiments are designed to compare inference-only techniques with fine-tuning approaches, providing a controlled environment to evaluate the trade-offs between distribution sharpening and task-reward-based optimization.

\textbf{Models}. We consider three models for our analysis: \texttt{Qwen2.5-3B-Instruct}, \texttt{Llama-3.2-3B-Instruct}, and the more recent \texttt{Qwen3-4B-Instruct-2507}.

\textbf{Training Datasets}. We focus on improving the mathematical reasoning capabilities of the mentioned LLMs, and thus use the Hendrycks math dataset \citep{hendrycks2021measuring} for fine-tuning the 3B models and the DeepScaleR dataset \citep{luo2025deepscaler} for the 4B model.

\textbf{Evaluation Datasets}. We evaluate the 3B models on Math-500 \citep{lightman2023let} and Minerva-Math \citep{lewkowycz2022solving} and the 4B model on the more challenging AIME 2024, AIME 2025 and HMMT 2025 datasets \citep{balunovic2025matharena}.

\textbf{Training Details}. We use a response length of $2048$ for training on Hendrycks math and $4096$ for DeepScaleR using the NeMo RL codebase \citep{nemo-rl}. We train all the models fully on-policy for $500$ and $1000$ steps for the 3B and 4B models respectively, and use the leave-one-out estimator (RLOO) to reduce variance \citep{ahmadian2024back}. We refer to \Cref{apdx:experiments} for further details about our implementation.

Beyond training, we investigate inference-time distribution sharpening through power sampling \citep{karan2025reasoning} and beam search \citep{chen2018stableeffectivelearningstrategy}. Through our experiments, we answer some of the key fundamental questions below.

\begin{figure}[t] 
\centering 
\includegraphics[width=\linewidth]{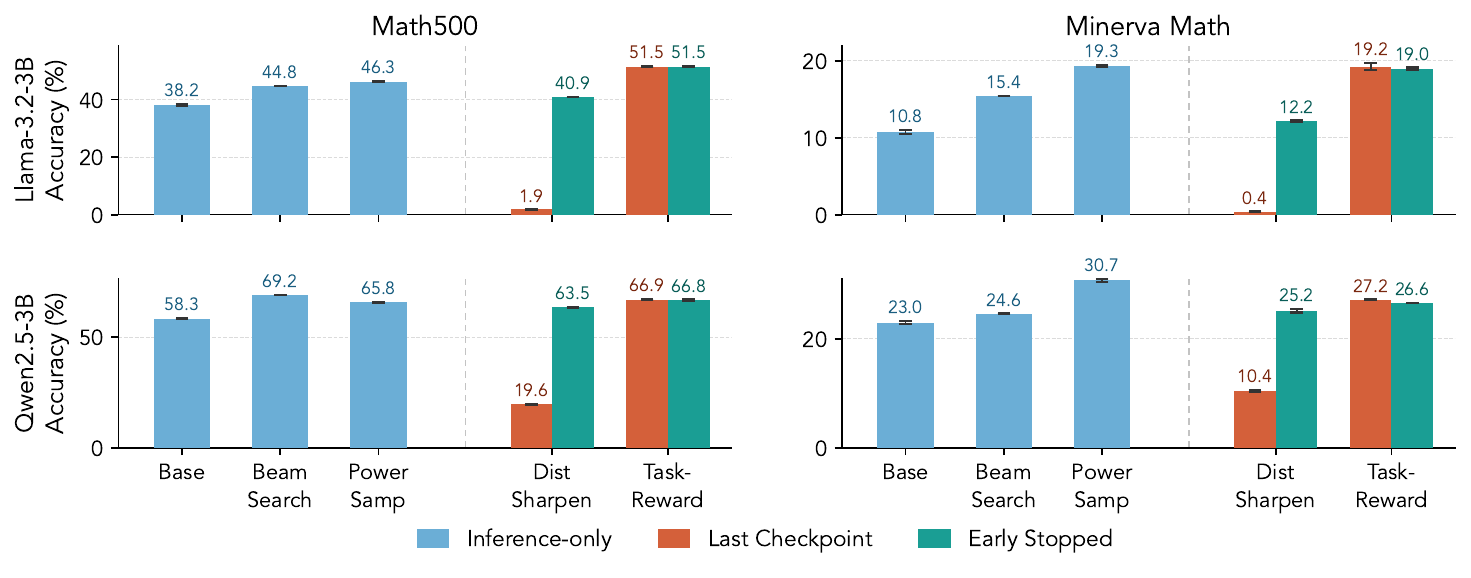} 
\vspace{-9mm}
\caption{\textbf{Pass@1 Accuracy of 3B models}: We compare inference-time distribution sharpening methods, task-reward RL and distribution sharpening based RL, where fine-tuning methods are evaluated at both the last and early stopped checkpoint. We observe that inference-time distribution sharpening can be competitive with task-reward based RL, while distribution sharpening based RL is highly unstable.} 
\vspace{-4mm}
\label{fig:math_3b} 
\end{figure}
\subsection{What do you gain from sharpening?}
We first experiment on smaller scale with \texttt{Llama-3.2-3B-Instruct} and \texttt{Qwen2.5-3B-Instruct} on Math-500 and Minerva-Math. \Cref{fig:math_3b} highlights that inference-time techniques like beam search and power sampling consistently improve over the base model and show the benefits of distribution sharpening as an inference time method. On the other hand, optimization-based sharpening of the distribution using RL initially improves performance but is eventually unstable and collapses: see the difference between early stopped and last checkpoints. In contrast, RL finetuning to optimize solely for task-reward is both stable and delivers good performance. It is important to note that beam search, power sampling, and early stopped RL based distribution sharpening are sometimes competitive, with inference-time methods being more consistent. 

\begin{figure}
    \centering
    \includegraphics[width=\linewidth]{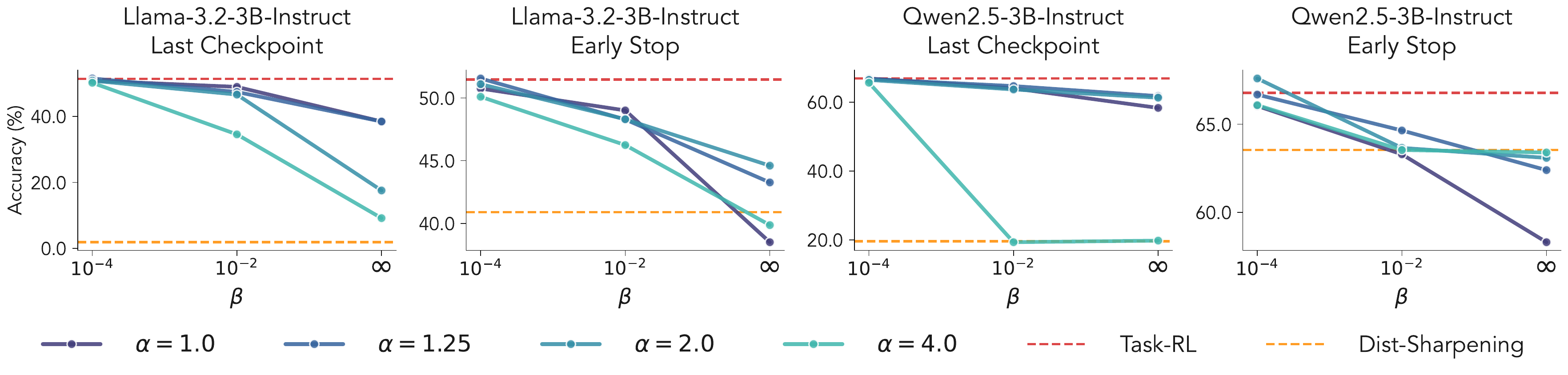}
    \vspace{-8mm}
    \caption{\textbf{Reliance on task-reward signal on Math-500}: For the tilted sampling optima, with $\beta = \infty$ denoting the tempered sampling optima, we see that increased reliance on the task-reward, i.e. lower $\beta$, leads to both improved as well as more stable performance.} 
    \vspace{-5mm}
    \label{fig:math-ab}
\end{figure}

We see similar instabilities when we finetune for tempered sampling, which is highlighted in \Cref{fig:math-ab} where runs corresponding to higher values of $\alpha$ for $\beta = \infty$ optimize for a tempered sampling optima and we see stark difference in performance between early stopped and final checkpoint across all tasks. 

\begin{figure}[t] 
\centering 
\includegraphics[width=\linewidth]{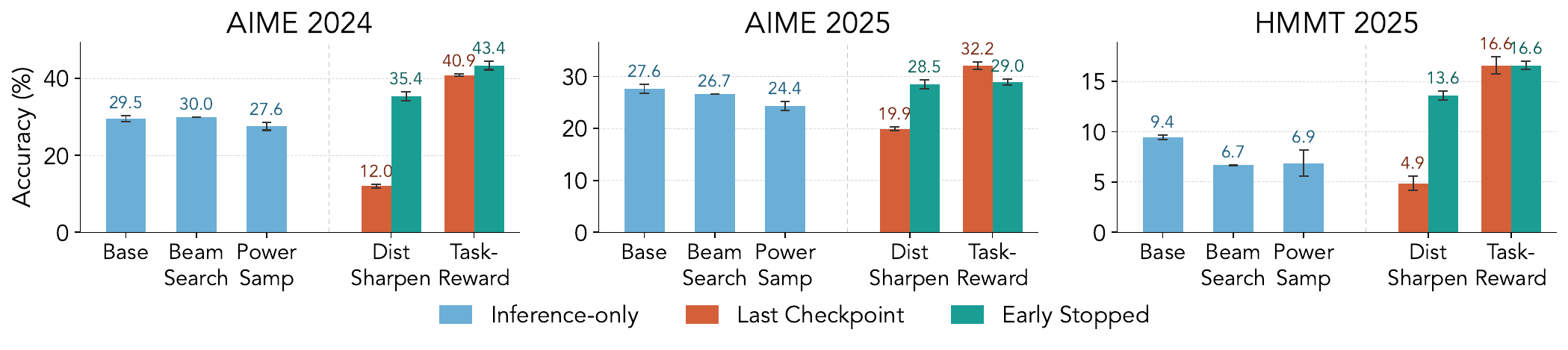} 
\vspace{-8mm}
\caption{\textbf{Pass@1 accuracy of 4B models}: We compare inference-time distribution sharpening methods, task-reward RL and distribution sharpening based RL, where fine-tuning methods are evaluated at both the last and early stopped checkpoint. We observe that task-reward based RL is consistently superior, with distribution sharpening RL being unstable but still superior to inference-only methods if early stopped. In fact, inference-time distribution sharpening methods are even inferior to standard ancestral sampling.} 
\vspace{-3mm}
\label{fig:math_4b} 
\end{figure}

Next, we turn our attention to harder tasks like AIME-24, AIME-25 and HMMT-25 with a stronger base model \texttt{Qwen3-4B-Instruct-2507}. In a similar analysis as above, we find that inference-only techniques geared at distribution sharpening -- both beam search and power sampling -- are much more sub-optimal even when compared to standard sampling (\Cref{fig:math_4b}). In fact, we find that leveraging RL to sharpen the distribution is considerably better provided it is managed through early stopping as the aforementioned instabilities still remain. Similar instabilities and trends for tempered sampling remain, as highlighted in \Cref{fig:qwen3-ab}. Finally, when we compare this to task-reward-based RL finetuning we again see consistently better performance as well as stable training, outperforming all the different forms of distribution sharpening. In general, these improvements remain even when considering tilted sampling, where smaller $\beta$ indicates larger tilting factor. Our experiments demonstrate a key finding:

\begin{observation}{1}
Distribution sharpening can sometimes bridge the performance gap to task-reward-based RL finetuning but it is often unreliable and unpredictable, while the latter delivers consistent and robust improvements.
\end{observation}

%

\subsection{Why is distribution sharpening unstable?}
The key reason behind the instability of distribution sharpening can be tracked down to first principles of how we handle generation of variable number of tokens. The predominant approach to handling variable response length relies on defining an alternative distribution 
\begin{align}
    \pi_\text{samp}(\rvy_{t+1} | \rvx, \rvy_{1:t}) = \begin{cases}
        \mathbb{1}_{\rvy_{t+1} = \texttt{EOS}} & \rvy_t = \texttt{EOS} \\
        \pi_\theta(\rvy_{t+1} | \rvx, \rvy_{1:t}) & \text{otherwise}
    \end{cases}
\end{align}
where the special token \texttt{EOS} denotes an absorbing state such that no new token is emitted once an \texttt{EOS} token is observed. In simpler terms, this leads to decoding tokens until either an \texttt{EOS} token is generated or the max token limit is achieved. Similarly log-likelihood calculation for a sequence also halts at the max token limit or at the first observation of \texttt{EOS}.

\begin{table}[t]
  \centering
  \footnotesize 
  \renewcommand{\arraystretch}{0.9} 
  \begin{tabular*}{\linewidth}{@{\extracolsep{\fill}}lllcccc@{}}
  \toprule
  & & & \multicolumn{2}{c}{Last Checkpoint} & \multicolumn{2}{c}{Early Stopped} \\
  \cmidrule(lr){4-5} \cmidrule(lr){6-7}
  Model Name & Mode & Length & Math500 & Minerva & Math500 & Minerva \\
  \midrule
  \multirow{4}{*}{Llama-3.2-3B-Instruct} 
  & \multirow{2}{*}{Dist. Sharpening} & Fixed & 44.5 & 18.3 & 45.5 & 18.1 \\
  & & Variable & {\color{red}1.9} & {\color{red}0.4} & 40.9 & 12.2 \\
  \cmidrule{2-7}
  & \multirow{2}{*}{Task-Reward} & Fixed & 50.6 & 17.3 & 50.7 & 17.4 \\
  & & Variable & \highlight\textbf{51.5} & \highlight\textbf{19.2} & \highlight\textbf{51.5} & \highlight\textbf{19.0} \\
  \midrule
  \multirow{4}{*}{Qwen2.5-3B-Instruct} 
  & \multirow{2}{*}{Dist. Sharpening} & Fixed & 60.0 & 24.6 & 63.1 & 26.1 \\
  & & Variable & {\color{red}19.6} & {\color{red}10.4} & 63.5 & 25.2 \\
  \cmidrule{2-7}
  & \multirow{2}{*}{Task-Reward} & Fixed & 65.8 & 26.7 & 65.8 & \highlight\textbf{26.9} \\
  & & Variable & \highlight\textbf{66.9} & \highlight\textbf{27.2} & \highlight\textbf{66.8} & 26.6 \\
  \bottomrule
  \end{tabular*}
  \vspace{-3mm}
  \caption{\textbf{Impact of fixed and variable length}: To analyze the instability of distribution sharpening, we experiment with a fixed length variant which ignores the \texttt{EOS} token in likelihood computation. We see that high likelihood samples are biased towards shorter sequences, and thus distribution sharpening while maintaining the capability of variable sequence lengths leads to unstable behavior -- shown in {\color{red}red}.}
  \vspace{-6mm}
  \label{tab:fixed_variable}
\end{table}
Since $\log \pi_\text{samp}(\rvy_{t+1} | \rvx, \rvy_{1:t}) \leq 0$, distribution sharpening through \Cref{eq:dist-sharpen,eq:tempered} naturally favors shorter responses as they lead to higher log likelihood. This phenomena has been previously widely studied in approaches like beam search \citep{murray-chiang-2018-correcting}, and it remains the core reason behind why such an optima is not favorable.

While several works address this issue by introducing length penalties, these solutions typically rely on heuristics \citep{yang-etal-2018-breaking}. Consequently, determining the optimal parameters for a given context remains a difficult and non-trivial task.

\looseness=-1
We validate this by RL finetuning with fixed response length, i.e. not treating \texttt{EOS} as a special absorption state for log-prob calculation, both for task-reward-based optimization as well as distribution sharpening. Evaluation is still done until the first utterance of \texttt{EOS} token. We see in \Cref{tab:fixed_variable} that unlike variable length training, distribution sharpening based RL on fixed length is stable and delivers improved performance over the base model that we saw earlier. We see a similar trend in both tilted and tempered sampling (\Cref{fig:fixedmath-ab}). However, such a framework is not preferred as we typically want such sequence models to define distributions over variable number of tokens. This clearly highlights the following key observation:

\begin{observation}{2}
The optima of distribution sharpening is inherently unfavorable. While some benefits can be gained from heuristics, early stopping,  and inference methods, better optimization of the underlying objective does not systematically imply improved performance or generalization.
\end{observation}

\subsection{Is RL to fault or distribution sharpening?}
A key reason behind both the collapse and limited performance improvement of distribution sharpening based finetuning could be the tool that we use: RL. In particular, this could be due to the objective being messy to optimize and would reflect in either training instabilities or the training reward not smoothly improving.

\begin{figure}[t]
    \centering
    \includegraphics[width=\linewidth]{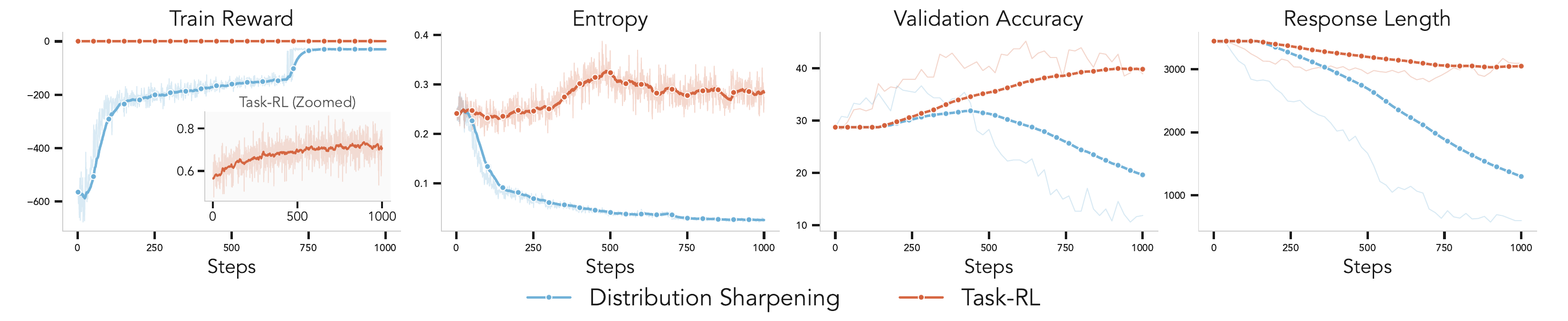}
    \vspace{-8mm}
    \caption{\textbf{Training Health of} \texttt{Qwen3-4B-Instruct-2507}: We monitor train reward, entropy, validation accuracy and response length when fine-tuning the 4B model for either task-reward or distribution sharpening based RL. Monotonic increase in train reward highlights that the learned policy is consistently improving in terms of the optimization objective.}
    \vspace{-4mm}
    \label{fig:train-dynamics}
\end{figure}

We observe the training dynamics of distribution sharpening and task-reward-based RL finetuning in \Cref{fig:train-dynamics} and observe that as expected from the previous section, distribution sharpening leads to consistent decrease in response length and entropy. We also observe an initial increase in validation performance followed by a sharp decline, highlighting again the importance of early stopping when relying on variable length formulation. Most importantly, we observe a consistent improvement in training reward which clearly showcases that the observations made in this study are not an artifact of poor or unstable optimization -- in fact RL finetuning consistently improves training reward and it is the optima itself that is not favorable. Based on this analysis, we make the following observation:

\begin{figure}[t]
    \centering
    \includegraphics[width=\linewidth]{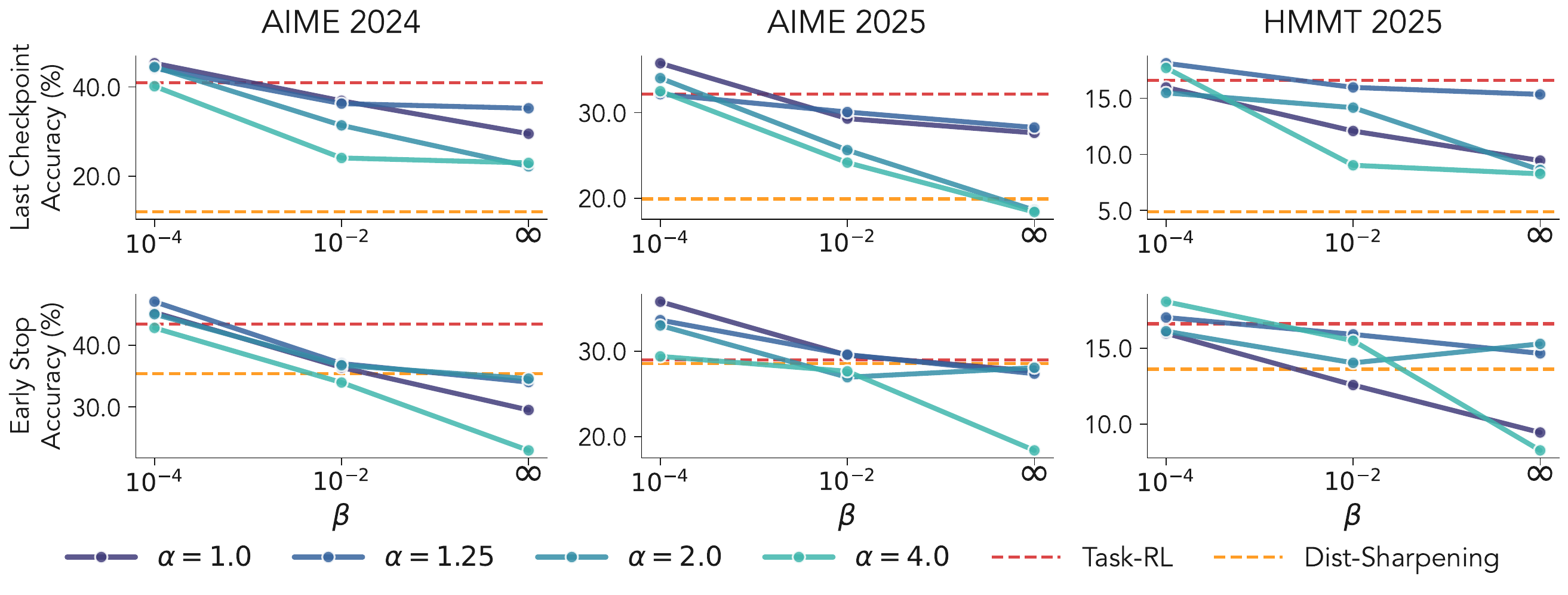}
    \vspace{-7mm}
    \caption{\textbf{Reliance on task-reward signal for }\texttt{Qwen3-4B-Instruct-2507}: For tilted sampling, with $\beta = \infty$ denoting the tempered sampling optima, we see that increased reliance on the task-reward, i.e. lower $\beta$, leads to both improved as well as more stable performance.}
    \vspace{-5mm}
    \label{fig:qwen3-ab}
\end{figure}
\begin{observation}{3}
The primary benefits of inference-time sharpening arise from limited compute which cannot fully optimize the objective. Employing more efficient optimization techniques like RL reveals inherent instabilities characteristic of the optima itself.
\end{observation}

\subsection{The importance of task-reward}
In the previous sections, we have already highlighted that while different forms of distribution sharpening -- inference-only, finetuning based sharpening and tempered sampling -- can sometimes improve performance, it is clear that when using task-reward signal for optimization we see very clear and consistent improvements in performance that are both stable and reliable.

To further validate this hypothesis, we consider the optima for tilted sampling in \Cref{eq:tilted}. Here, we can clearly see that $\alpha$ plays the role of sharpening the distribution paired by a tilting factor, where $\beta$ governs the impact of task-reward on training.

\begin{figure}
    \centering
    \includegraphics[width=\linewidth]{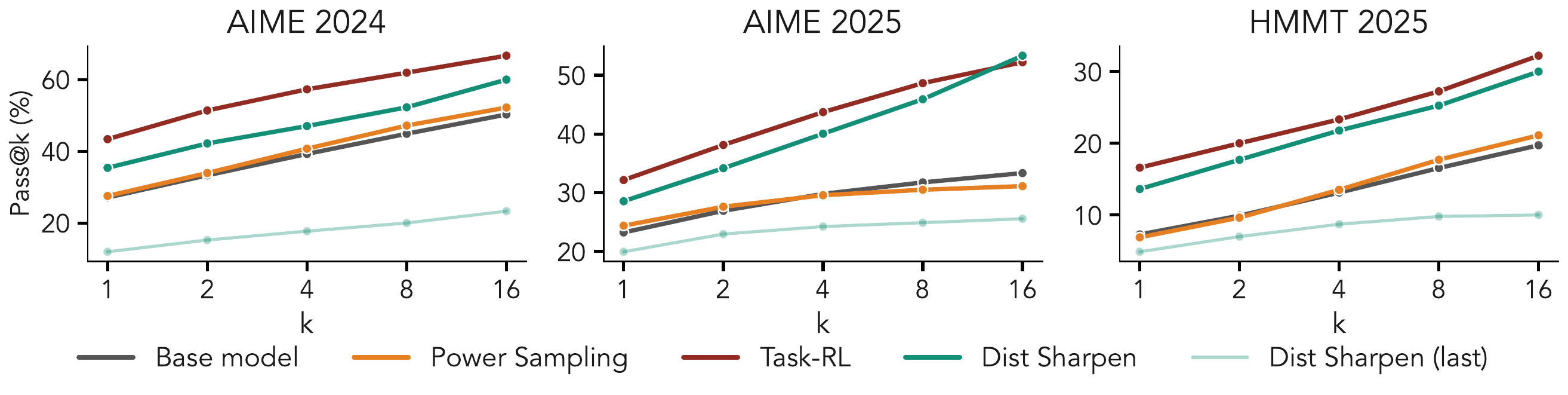}
    \vspace{-8.5mm}
    \caption{\textbf{Pass@k performance with} \texttt{Qwen3-4B-Instruct-2507}: We see that task-reward maximization also leads to improved Pass@k performance for various k. We also see that distribution sharpening based RL, while inferior to Task-RL, is better than the base model. However, it is unstable and eventually collapses -- see Dist Sharpen (last). Finally, we note that Power Sampling does not differ much from the base model.}
    \vspace{-6,5mm}
    \label{fig:pass_k}
\end{figure}

Across all our experiments in \Cref{fig:math-ab,fig:fixedmath-ab,fig:qwen3-ab} we see a very similar trend where an increasing reliance on task-reward (reducing $\beta$) not only improves the final performance of the model but also provides additional stability when training with variable response lengths, with the extreme of removing task-reward ($\beta = \infty$) yielding tempered sampling which proves to be highly unstable and not performative. Finally, we also observe an improvement in Pass@k performance with task-reward based optimization, superior to the unstable distribution sharpening fine-tuning (\Cref{fig:pass_k}). Thus, we observe:

\begin{observation}{4}
Keeping everything else constant, increasing the reliance on task-reward in the optima improves performance and stability over tempered sampling.
\end{observation}

\subsection{To sharpen or to reward?}
This brings us to a pivotal question of whether we should fine-tune LLMs for distribution sharpening or optimizing task-reward? We have already established that relying strongly on the task-reward is pivotal for improved performance and stable training, but we have not yet discussed if this can be combined with tempered sampling. Luckily, tilted sampling provides us with a control to study this: $\beta$ governs the reliance on task-reward and $\alpha$ the amount of distribution sharpening. This lever provides us a way of combining environment signal and inherent exploration, where larger $\alpha$ biases the model towards less exploration and smaller $\beta$ incentivizes it to prioritize higher task rewards. 

We observe that for easier tasks (\Cref{fig:math-ab,fig:fixedmath-ab}) solely relying on task-reward is the best strategy (compare dashed red with solid lines). However, for more complex problems in \Cref{fig:qwen3-ab}, we sometimes do see that small amounts of sharpening $\alpha = 1.25$ can sometimes lead to improved performance; though the difference is marginal. Thus, we observe

\begin{observation}{5}
Leveraging both distribution sharpening and task-reward in tilted sampling only sometimes provides marginal improvements, but is unstable for high $\beta$ or high $\alpha$.
\end{observation}

\section{Conclusion}
\vspace{-2mm}
We present a unified comparison between distribution sharpening and task-reward-based fine-tuning within a KL-regularized reinforcement learning framework. Our results show that while distribution sharpening can yield modest gains, it is fundamentally limited by unfavorable optima and instability, particularly in variable-length generation. In contrast, task-reward optimization consistently delivers stable and significant improvements, especially on harder and more complex tasks where the base model is uncertain.

Overall, our findings suggest that the benefits of task-reward-based fine-tuning cannot be reduced to distribution sharpening alone. Instead, task-dependent reward signals play a central role in shaping model behavior and enabling capability improvements. Within a shared training paradigm, this is showcased by consistent improvement in performance with increased reliance on task-reward, i.e. lower $\beta$. 

While combining sharpening with task-reward can sometimes offer marginal gains in some settings, these effects are neither robust nor necessary, reinforcing the importance of task-reward design in post-training. We believe that conducting similar analysis over longer sequence length training as well as a mixture of different tasks would be an important future work to further highlight the importance of task-rewards over distribution sharpening.


\section*{Acknowledgments}
The research was enabled in part by computational resources provided by the Digital Research
Alliance of Canada (\url{https://alliancecan.ca}) and Mila (\url{https://mila.quebec}).

GL acknowledges support from NSERC Discovery Grant RGPIN-2018-04821, the Canada Research Chair in Neural Computations and Interfacing, and a Canada-CIFAR AI Chair. SM acknowledges funding from FRQNT Doctoral Fellowship (\url{https://doi.org/10.69777/372208}).

\clearpage

\bibliography{colm2026_conference}
\bibliographystyle{colm2026_conference}

\clearpage
\appendix
\section*{Appendix}

\section{Implementation Details}
\label{apdx:experiments}
In this section, we provide additional details about the experiments conducted for both fine-tuning as well as inference-time methods.

\textbf{RL Fine-tuning}: We use a batch size of $32$ different problems for the 3B models and $64$ for the 4B models. Each model is trained with a group size of $8$ and uses the leave-one-out estimator for variance reduction \citep{ahmadian2024back}. The 3B models are trained with a response length of $2048$ for $500$ steps while the 4B models are trained with a response length of $4096$ for $1000$ steps. We train using a learning rate of $10^{-6}$ with a linear warmup over $50$ steps from an initial rate of $10^{-7}$, using the \texttt{AdamW} \citep{loshchilov2017decoupled} optimizer with $0.01$ weight decay.

While it is much more efficient to leverage asynchronous RL training where responses can be generated slightly off-policy from an earlier iteration model \citep{mnih2016asynchronousmethodsdeepreinforcement}, we limit our study to the synchronous setting since the former can introduce bias and instability \citep{munos2016safeefficientoffpolicyreinforcement}. Even with synchronous training, we leverage importance sampling to correct for the slight off-policyness suffered from the trainer / inference mismatch \citep{precup2000eligibility}.

For evaluation on all RL fine-tuned models, we use top-p=1.0, temperature=1.0, and top-k=1.0, since this is the sampling setup used for the on-policy sampling in RL.

\textbf{Power Sampling}:
We took the vLLM implementation of \cite{zuo2024mhllm}, ensured it was in line with the original implementation (fixed some bugs) and adapted it to work with Nemo-RL evaluation pipeline. We use default hyper-parameters from the original paper:
\begin{itemize}
    \item Block size = Sequence length/16
    \item MCMC steps = 10
\end{itemize}
\textbf{Beam Search}: 
We use vLLM beam search implementation and the following hyperparamters:
\begin{itemize}
    \item 4 beams
    \item Length penalty = 0.0
    \item Top-p : 1.0
    \item Temperature : 1.0
    \item Top-k : -1
\end{itemize}

\begin{wrapfigure}{r}{0.5\textwidth}
  \centering
  \vspace{-1cm}
  \includegraphics[width=0.48\textwidth]{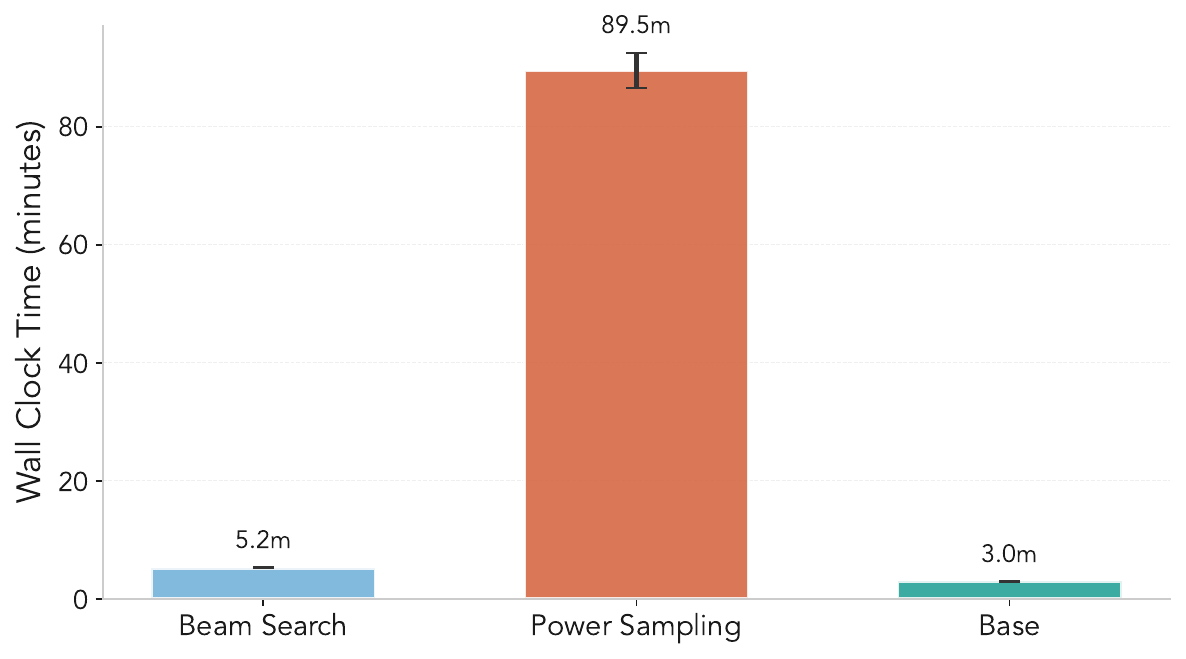}
  \vspace{-3mm}
  \caption{\textbf{Inference Time Comparison}: We compare the time taken for performing inference with different inference-time methods.}
  \label{fig:time}
\end{wrapfigure}
\textbf{Inference Time}: We evaluate the amount of time taken to perform inference using the different inference-time procedures. We see that standard sampling with vLLM is extremely fast, with beam search using only a little more computation overhead. However, power sampling \citep{karan2025reasoning} which relies on an expensive MCMC procedure is considerably more expensive. \Cref{fig:time} highlights this difference in efficiency during inference and shows that MCMC based approaches are predominantly ill-disposed for efficient inference.

\section{Fixed Response Length RL}
To better analyze the limitations of distribution sharpening based RL fine-tuning, we study a specific case where we always generate responses up to a fixed response length -- irrespective of whether an \texttt{EOS} token is observed. This rollout is then used to perform RL fine-tuning, where log-probabilities are computed based on the entire fixed response.

\begin{figure}[t]
    \centering
    \includegraphics[width=\linewidth]{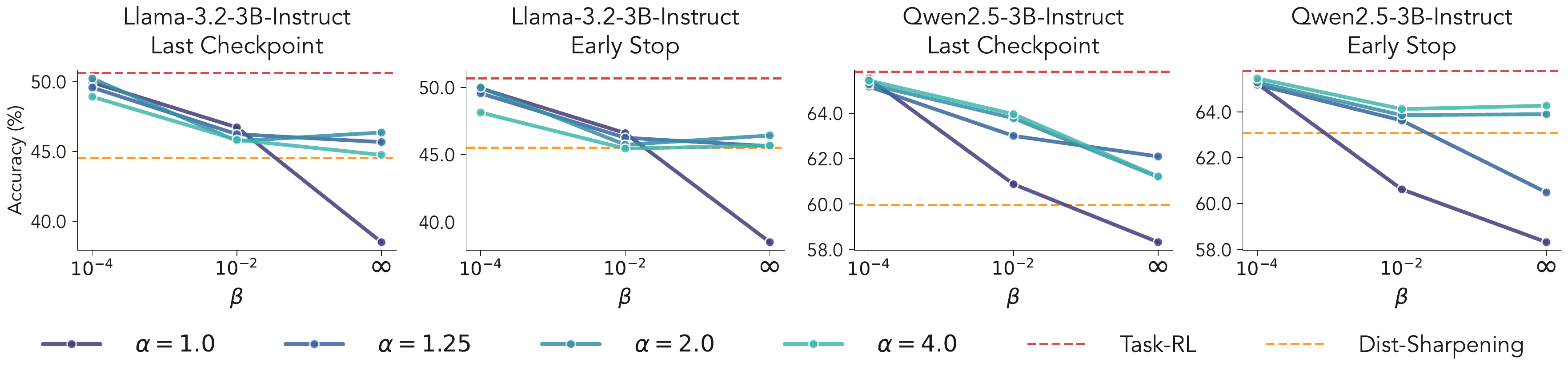}
    \vspace{-8mm}
    \caption{\textbf{Reliance on task-reward signal on Math-500 for models trained with fixed length}: We see that increased reliance on task-reward, i.e. lower $\beta$, leads to both improved as well as more stable performance.}
    \vspace{-6mm}
    \label{fig:fixedmath-ab}
\end{figure}

\Cref{fig:fixedmath-ab} highlights the performance corresponding to both task-reward maximization as well as optimization based distribution sharpening. It also highlights tempered and tilted sampling, showing that it is indeed the ability of sampling variable response length that leads to instabilities in distribution sharpening.

\clearpage
\section{Full evaluation results of trained models}

\begin{table}[t]
\centering
\footnotesize
\caption{Qwen2.5-3B - Variable, Last checkpoint}
\begin{tabular}{lcccccccc}
\toprule
& & & \multicolumn{3}{c}{Math-500} & \multicolumn{3}{c}{Minerva} \\
Strategy & $\alpha$ & $\beta$ & Pass@1 & Pass@16 & Maj@16 & Pass@1 & Pass@16 & Maj@16 \\
\midrule
Base & -- & -- & 58.3 & 87.5 & 66.3 & 23.0 & 47.8 & 27.2 \\
\multirow{3}{*}{MH-LLM} & 1.0 & -- & 47.7 & 82.9 & 62.5 & 20.6 & 46.8 & 25.6 \\
 & 2.0 & -- & 64.8 & 87.1 & 71.1 & 29.1 & 43.8 & 31.7 \\
 & 4.0 & -- & 65.8 & 85.5 & 71.1 & \highlight \textbf{30.7} & 44.2 & \highlight \textbf{32.2} \\
Beam Search & -- & -- & \highlight \textbf{69.2} & -- & -- & 24.6 & -- & -- \\
\midrule
Task-Reward & -- & -- & 66.9 & 87.9 & \highlight \textbf{72.1} & 27.2 & 47.4 & 29.4 \\
Dist. Sharpen & -- & -- & \textcolor{red}{19.6} & \textcolor{red}{26.3} & \textcolor{red}{19.5} & \textcolor{red}{10.4} & \textcolor{red}{15.9} & \textcolor{red}{10.7} \\
\midrule
\multirow{12}{*}{Tilted Sampling} & \multirow{3}{*}{1.00} & $\infty$ & 58.3 & 87.5 & 66.3 & 23.0 & 47.8 & 27.2 \\
 &  & $10^{-2}$ & 64.0 & \highlight \textbf{88.7} & 70.0 & 25.7 & \highlight \textbf{48.9} & 27.3 \\
 &  & $10^{-4}$ & 66.9 & 88.2 & 71.5 & 26.7 & 46.8 & 28.2 \\
\cmidrule{2-9}
 & \multirow{3}{*}{1.25} & $\infty$ & 61.7 & 87.1 & 68.1 & 25.6 & 46.8 & 28.9 \\
 &  & $10^{-2}$ & 64.6 & 88.2 & 70.3 & 26.9 & 48.7 & 28.7 \\
 &  & $10^{-4}$ & 66.7 & 88.6 & 71.9 & 26.6 & 47.1 & 28.8 \\
\cmidrule{2-9}
 & \multirow{3}{*}{2.00} & $\infty$ & 61.3 & 84.7 & 65.7 & 26.5 & 45.0 & 28.6 \\
 &  & $10^{-2}$ & 63.7 & 87.0 & 68.5 & 26.4 & 46.7 & 28.7 \\
 &  & $10^{-4}$ & 66.4 & 87.9 & 70.7 & 26.5 & 47.2 & 28.3 \\
\cmidrule{2-9}
 & \multirow{3}{*}{4.00} & $\infty$ & \textcolor{red}{19.8} & \textcolor{red}{30.1} & \textcolor{red}{19.9} & \textcolor{red}{10.8} & \textcolor{red}{18.9} & \textcolor{red}{11.0} \\
 &  & $10^{-2}$ & \textcolor{red}{19.4} & \textcolor{red}{30.3} & \textcolor{red}{19.2} & \textcolor{red}{10.5} & \textcolor{red}{16.2} & \textcolor{red}{10.9} \\
 &  & $10^{-4}$ & 65.6 & 87.9 & 70.9 & 25.9 & 46.7 & 28.3 \\
\bottomrule
\end{tabular}
\end{table}

\begin{table}[t]
\centering
\footnotesize
\caption{Qwen2.5-3B - Variable, Best checkpoint}
\begin{tabular}{lcccccccc}
\toprule
& & & \multicolumn{3}{c}{Math-500} & \multicolumn{3}{c}{Minerva} \\
Strategy & $\alpha$ & $\beta$ & Pass@1 & Pass@16 & Maj@16 & Pass@1 & Pass@16 & Maj@16 \\
\midrule
Base & -- & -- & 58.3 & 87.5 & 66.3 & 23.0 & 47.8 & 27.2 \\
\multirow{3}{*}{MH-LLM} & 1.0 & -- & {47.7} & {82.9} & {62.5} & {20.6} & {46.8} & {25.6} \\
 & 2.0 & -- & 64.8 & 87.1 & 71.1 & 29.1 & 43.8 & 31.7 \\
 & 4.0 & -- & 65.8 & 85.5 & 71.1 & \highlight \textbf{30.7} & 44.2 & \highlight \textbf{32.2} \\
Beam Search & -- & -- & \highlight \textbf{69.2} & -- & -- & 24.6 & -- & -- \\
\midrule
Task-Reward & -- & -- & 66.8 & 87.8 & 72.2 & 26.6 & 46.4 & 29.0 \\
Dist. Sharpen & -- & -- & 63.5 & \highlight \textbf{88.9} & 69.1 & 25.2 & 47.5 & 28.2 \\
\midrule
\multirow{12}{*}{Tilted Sampling} & \multirow{3}{*}{1.00} & $\infty$ & 58.3 & 87.5 & 66.3 & 23.0 & 47.8 & 27.2 \\
 &  & $10^{-2}$ & 63.3 & 88.7 & 70.1 & 25.2 & 48.2 & 25.5 \\
 &  & $10^{-4}$ & 66.0 & 88.6 & 71.0 & 26.8 & 47.3 & 28.6 \\
\cmidrule{2-9}
 & \multirow{3}{*}{1.25} & $\infty$ & 62.4 & \highlight \textbf{88.9} & 69.1 & 24.7 & 48.5 & 26.8 \\
 &  & $10^{-2}$ & 64.6 & 88.2 & 70.3 & 26.9 & 48.7 & 28.7 \\
 &  & $10^{-4}$ & 66.7 & 88.6 & 71.9 & 26.6 & 47.1 & 28.8 \\
\cmidrule{2-9}
 & \multirow{3}{*}{2.00} & $\infty$ & 63.1 & 88.1 & 68.6 & 25.6 & 47.5 & 29.2 \\
 &  & $10^{-2}$ & 63.7 & 87.0 & 68.5 & 26.4 & 46.7 & 28.7 \\
 &  & $10^{-4}$ & 67.6 & 88.4 & \highlight \textbf{72.5} & 27.1 & \highlight \textbf{49.0} & 29.5 \\
\cmidrule{2-9}
 & \multirow{3}{*}{4.00} & $\infty$ & 63.4 & 88.5 & 70.1 & 25.0 & 47.5 & 27.1 \\
 &  & $10^{-2}$ & 63.5 & \highlight \textbf{88.9} & 69.9 & 25.8 & 47.2 & 28.3 \\
 &  & $10^{-4}$ & 66.1 & 88.9 & 70.9 & 26.1 & 47.4 & 26.8 \\
\bottomrule
\end{tabular}
\end{table}

\begin{table}[t]
\centering
\footnotesize
\caption{Qwen2.5-3B - Fixed, Last checkpoint}
\begin{tabular}{lcccccccc}
\toprule
& & & \multicolumn{3}{c}{Math-500} & \multicolumn{3}{c}{Minerva} \\
Strategy & $\alpha$ & $\beta$ & Pass@1 & Pass@16 & Maj@16 & Pass@1 & Pass@16 & Maj@16 \\
\midrule
Base & -- & -- & -- & -- & -- & -- & -- & -- \\
\multirow{3}{*}{MH-LLM} & 1.0 & -- & 47.7 & 82.9 & 62.5 & 20.6 & 46.8 & 25.6 \\
 & 2.0 & -- & 64.8 & 87.1 & 71.1 & 29.1 & 43.8 & 31.7 \\
 & 4.0 & -- & 65.8 & 85.5 & 71.1 & \highlight \textbf{30.7} & 44.2 & \highlight \textbf{32.2} \\
Beam Search & -- & -- & \highlight \textbf{69.2} & -- & -- & 24.6 & -- & -- \\
\midrule
Task-Reward & -- & -- & 65.8 & 89.4 & 70.8 & 26.7 & 46.2 & 29.4 \\
Dist. Sharpen & -- & -- & 60.0 & 79.6 & 63.5 & 24.6 & 43.1 & 26.5 \\
\midrule
\multirow{12}{*}{Tilted Sampling} & \multirow{3}{*}{1.00} & $\infty$ & 58.3 & 87.5 & 66.3 & 23.0 & 47.8 & 27.2 \\
 &  & $10^{-2}$ & 60.9 & 89.0 & 68.8 & 24.1 & 48.0 & 27.3 \\
 &  & $10^{-4}$ & 65.5 & 88.9 & 69.9 & 26.5 & 46.6 & 27.6 \\
\cmidrule{2-9}
 & \multirow{3}{*}{1.25} & $\infty$ & 62.1 & 88.3 & 69.3 & 26.0 & \highlight \textbf{49.1} & 28.7 \\
 &  & $10^{-2}$ & 63.0 & 88.5 & 69.9 & 25.7 & 47.2 & 28.7 \\
 &  & $10^{-4}$ & 65.2 & 88.5 & 70.9 & 25.7 & 48.4 & 27.2 \\
\cmidrule{2-9}
 & \multirow{3}{*}{2.00} & $\infty$ & 61.2 & 86.4 & 67.1 & 24.4 & 47.2 & 28.9 \\
 &  & $10^{-2}$ & 63.8 & 89.1 & 69.7 & 26.4 & 48.3 & 28.6 \\
 &  & $10^{-4}$ & 65.3 & \highlight \textbf{89.9} & 69.5 & 26.6 & 47.4 & 29.9 \\
\cmidrule{2-9}
 & \multirow{3}{*}{4.00} & $\infty$ & 61.2 & 84.2 & 64.9 & 24.3 & 45.0 & 26.7 \\
 &  & $10^{-2}$ & 64.0 & 87.7 & 68.5 & 24.1 & 46.0 & 26.3 \\
 &  & $10^{-4}$ & 65.5 & 88.9 & \highlight \textbf{71.2} & 25.3 & 46.3 & 28.1 \\
\bottomrule
\end{tabular}
\end{table}

\begin{table}[t]
\centering
\footnotesize
\caption{Qwen2.5-3B - Fixed, Best checkpoint}
\begin{tabular}{lcccccccc}
\toprule
& & & \multicolumn{3}{c}{Math-500} & \multicolumn{3}{c}{Minerva} \\
Strategy & $\alpha$ & $\beta$ & Pass@1 & Pass@16 & Maj@16 & Pass@1 & Pass@16 & Maj@16 \\
\midrule
Base & -- & -- & -- & -- & -- & -- & -- & -- \\
\multirow{3}{*}{MH-LLM} & 1.0 & -- & 47.7 & 82.9 & 62.5 & 20.6 & 46.8 & 25.6 \\
 & 2.0 & -- & 64.8 & 87.1 & 71.1 & 29.1 & 43.8 & 31.7 \\
 & 4.0 & -- & 65.8 & 85.5 & 71.1 & \highlight \textbf{30.7} & 44.2 & \highlight \textbf{32.2} \\
Beam Search & -- & -- & \highlight \textbf{69.2} & -- & -- & 24.6 & -- & -- \\
\midrule
Task-Reward & -- & -- & 65.8 & \highlight \textbf{90.4} & 70.7 & 26.9 & \highlight \textbf{49.0} & 28.2 \\
Dist. Sharpen & -- & -- & 63.1 & 87.2 & 68.9 & 26.1 & 47.9 & 28.7 \\
\midrule
\multirow{12}{*}{Tilted Sampling} & \multirow{3}{*}{1.00} & $\infty$ & 58.3 & 87.5 & 66.3 & 23.0 & 47.8 & 27.2 \\
 &  & $10^{-2}$ & 60.6 & 89.1 & 67.8 & 24.0 & 48.7 & 26.3 \\
 &  & $10^{-4}$ & 65.2 & 89.3 & 71.1 & 26.2 & 47.5 & 28.6 \\
\cmidrule{2-9}
 & \multirow{3}{*}{1.25} & $\infty$ & 60.5 & 88.9 & 68.9 & 23.6 & 48.0 & 26.1 \\
 &  & $10^{-2}$ & 63.6 & 89.3 & 69.7 & 25.7 & 46.8 & 27.6 \\
 &  & $10^{-4}$ & 65.2 & 88.5 & 70.9 & 25.7 & 48.4 & 27.2 \\
\cmidrule{2-9}
 & \multirow{3}{*}{2.00} & $\infty$ & 63.9 & 88.3 & 69.7 & 24.5 & 46.7 & 27.9 \\
 &  & $10^{-2}$ & 63.9 & 89.1 & 68.8 & 26.2 & 47.5 & 28.8 \\
 &  & $10^{-4}$ & 65.3 & 88.9 & 70.3 & 26.7 & 47.4 & 28.3 \\
\cmidrule{2-9}
 & \multirow{3}{*}{4.00} & $\infty$ & 64.3 & 89.8 & 70.8 & 24.7 & 46.9 & 27.7 \\
 &  & $10^{-2}$ & 64.1 & 90.0 & \highlight \textbf{71.3} & 25.0 & 47.7 & 26.8 \\
 &  & $10^{-4}$ & 65.5 & 88.1 & 70.0 & 25.3 & 47.2 & 26.2 \\
\bottomrule
\end{tabular}
\end{table}

\begin{table}[t]
\centering
\footnotesize
\caption{Llama-3.2-3B - Variable, Last checkpoint}
\begin{tabular}{lcccccccc}
\toprule
& & & \multicolumn{3}{c}{Math-500} & \multicolumn{3}{c}{Minerva} \\
Strategy & $\alpha$ & $\beta$ & Pass@1 & Pass@16 & Maj@16 & Pass@1 & Pass@16 & Maj@16 \\
\midrule
Base & -- & -- & 38.2 & 76.3 & 49.3 & 10.8 & 37.5 & 14.3 \\
\multirow{3}{*}{MH-LLM} & 1.0 & -- & 30.5 & 70.5 & 40.9 & 9.3 & 34.4 & 12.9 \\
 & 2.0 & -- & 44.4 & 74.1 & 50.1 & 18.2 & 41.3 & 21.1 \\
 & 4.0 & -- & 46.3 & 72.2 & 49.6 & 19.3 & 38.1 & 21.1 \\
Beam Search & -- & -- & 44.8 & -- & -- & 15.4 & -- & -- \\
\midrule
Task-Reward & -- & -- & 51.5 & 76.9 & 55.5 & 19.2 & 40.6 & 21.8 \\
Dist. Sharpen & -- & -- & \textcolor{red}{1.9} & \textcolor{red}{2.7} & \textcolor{red}{1.7} & \textcolor{red}{0.4} & \textcolor{red}{1.3} & \textcolor{red}{0.2} \\
\midrule
\multirow{12}{*}{Tilted Sampling} & \multirow{3}{*}{1.00} & $\infty$ & 38.5 & 78.0 & 48.6 & 11.4 & 38.0 & 14.0 \\
 &  & $10^{-2}$ & 49.0 & 78.1 & 56.6 & 17.6 & 41.1 & 21.6 \\
 &  & $10^{-4}$ & 50.9 & 78.5 & 55.9 & 19.5 & 42.8 & 21.3 \\
\cmidrule{2-9}
 & \multirow{3}{*}{1.25} & $\infty$ & 38.5 & 64.3 & 45.2 & 13.9 & 39.2 & 16.4 \\
 &  & $10^{-2}$ & 47.5 & \highlight \textbf{79.1} & 53.3 & 17.6 & 42.2 & 20.1 \\
 &  & $10^{-4}$ & \highlight \textbf{51.6} & 78.5 & \highlight \textbf{57.3} & 19.9 & \highlight \textbf{44.9} & 21.4 \\
\cmidrule{2-9}
 & \multirow{3}{*}{2.00} & $\infty$ & \textcolor{red}{17.6} & \textcolor{red}{30.4} & \textcolor{red}{19.0} & \textcolor{red}{5.3} & \textcolor{red}{21.1} & \textcolor{red}{5.9} \\
 &  & $10^{-2}$ & 46.7 & 69.5 & 51.5 & 18.4 & 41.7 & 20.3 \\
 &  & $10^{-4}$ & 50.9 & 78.6 & 55.8 & \highlight \textbf{20.1} & 42.3 & \highlight \textbf{22.4} \\
\cmidrule{2-9}
 & \multirow{3}{*}{4.00} & $\infty$ & \textcolor{red}{9.2} & \textcolor{red}{15.7} & \textcolor{red}{9.5} & \textcolor{red}{1.6} & \textcolor{red}{8.8} & \textcolor{red}{1.1} \\
 &  & $10^{-2}$ & 34.6 & 49.2 & 36.9 & 12.7 & 24.6 & 12.7 \\
 &  & $10^{-4}$ & 50.2 & 78.5 & 56.0 & 19.1 & 42.9 & 21.1 \\
\bottomrule
\end{tabular}
\end{table}

\begin{table}[t]
\centering
\footnotesize
\caption{Llama-3.2-3B - Variable, Best checkpoint}
\begin{tabular}{lcccccccc}
\toprule
& & & \multicolumn{3}{c}{Math-500} & \multicolumn{3}{c}{Minerva} \\
Strategy & $\alpha$ & $\beta$ & Pass@1 & Pass@16 & Maj@16 & Pass@1 & Pass@16 & Maj@16 \\
\midrule
Base & -- & -- & 38.2 & 76.3 & 49.3 & 10.8 & 37.5 & 14.3 \\
\multirow{3}{*}{MH-LLM} & 1.0 & -- & 30.5 & 70.5 & 40.9 & 9.3 & 34.4 & 12.9 \\
 & 2.0 & -- & 44.4 & 74.1 & 50.1 & 18.2 & 41.3 & 21.1 \\
 & 4.0 & -- & 46.3 & 72.2 & 49.6 & 19.3 & 38.1 & 21.1 \\
Beam Search & -- & -- & 44.8 & -- & -- & 15.4 & -- & -- \\
\midrule
Task-Reward & -- & -- & 51.5 & 79.5 & 57.0 & 19.0 & 42.3 & 21.1 \\
Dist. Sharpen & -- & -- & 40.9 & 79.5 & 49.6 & 12.2 & 39.1 & 14.7 \\
\midrule
\multirow{12}{*}{Tilted Sampling} & \multirow{3}{*}{1.00} & $\infty$ & 38.5 & 78.0 & 48.6 & 11.4 & 38.0 & 14.0 \\
 &  & $10^{-2}$ & 49.0 & 78.1 & 56.6 & 17.6 & 41.1 & 21.6 \\
 &  & $10^{-4}$ & 50.7 & 79.6 & 56.7 & 19.3 & 43.4 & 21.6 \\
\cmidrule{2-9}
 & \multirow{3}{*}{1.25} & $\infty$ & 43.3 & 79.1 & 50.6 & 15.4 & 41.3 & 18.0 \\
 &  & $10^{-2}$ & 48.3 & \highlight \textbf{80.4} & 54.1 & 18.3 & 44.5 & 21.0 \\
 &  & $10^{-4}$ & \highlight \textbf{51.6} & 78.5 & \highlight \textbf{57.3} & 19.9 & \highlight \textbf{44.9} & 21.4 \\
\cmidrule{2-9}
 & \multirow{3}{*}{2.00} & $\infty$ & 44.6 & 78.6 & 52.3 & 16.1 & 41.3 & 18.1 \\
 &  & $10^{-2}$ & 48.3 & 75.6 & 54.2 & 18.8 & 43.8 & 21.2 \\
 &  & $10^{-4}$ & 51.1 & 78.5 & 56.1 & \highlight \textbf{20.3} & 43.4 & \highlight \textbf{22.7} \\
\cmidrule{2-9}
 & \multirow{3}{*}{4.00} & $\infty$ & 39.9 & 78.2 & 49.7 & 11.7 & 39.0 & 14.6 \\
 &  & $10^{-2}$ & 46.2 & 76.5 & 52.1 & 18.3 & 42.5 & 20.6 \\
 &  & $10^{-4}$ & 50.1 & 79.2 & 55.5 & 18.9 & 44.5 & 20.2 \\
\bottomrule
\end{tabular}
\end{table}

\begin{table}[t]
\centering
\footnotesize
\caption{Llama-3.2-3B - Fixed, Last checkpoint}
\begin{tabular}{lcccccccc}
\toprule
& & & \multicolumn{3}{c}{Math-500} & \multicolumn{3}{c}{Minerva} \\
Strategy & $\alpha$ & $\beta$ & Pass@1 & Pass@16 & Maj@16 & Pass@1 & Pass@16 & Maj@16 \\
\midrule
Base & -- & -- & -- & -- & -- & -- & -- & -- \\
\multirow{3}{*}{MH-LLM} & 1.0 & -- & 30.5 & 70.5 & 40.9 & 9.3 & 34.4 & 12.9 \\
 & 2.0 & -- & 44.4 & 74.1 & 50.1 & 18.2 & 41.3 & 21.1 \\
 & 4.0 & -- & 46.3 & 72.2 & 49.6 & \highlight \textbf{19.3} & 38.1 & 21.1 \\
Beam Search & -- & -- & 44.8 & -- & -- & 15.4 & -- & -- \\
\midrule
Task-Reward & -- & -- & \highlight \textbf{50.6} & 79.2 & 55.5 & 17.3 & 41.9 & 18.9 \\
Dist. Sharpen & -- & -- & 44.5 & 72.4 & 48.2 & 18.3 & 39.0 & 20.1 \\
\midrule
\multirow{12}{*}{Tilted Sampling} & \multirow{3}{*}{1.00} & $\infty$ & 38.5 & 78.0 & 48.6 & 11.4 & 38.0 & 14.0 \\
 &  & $10^{-2}$ & 46.7 & 78.9 & 54.7 & 16.3 & \highlight \textbf{43.4} & 18.5 \\
 &  & $10^{-4}$ & 49.9 & 78.1 & 55.3 & 18.7 & 42.4 & 21.9 \\
\cmidrule{2-9}
 & \multirow{3}{*}{1.25} & $\infty$ & 45.7 & 79.9 & 52.3 & 17.1 & 43.1 & 20.0 \\
 &  & $10^{-2}$ & 46.2 & \highlight \textbf{80.4} & 54.1 & 16.3 & 41.7 & 19.4 \\
 &  & $10^{-4}$ & 49.6 & 78.5 & 55.3 & 19.2 & 42.8 & 20.8 \\
\cmidrule{2-9}
 & \multirow{3}{*}{2.00} & $\infty$ & 46.4 & 77.5 & 51.9 & 18.0 & 41.9 & 20.0 \\
 &  & $10^{-2}$ & 45.7 & 78.3 & 52.8 & 17.2 & 42.3 & 20.2 \\
 &  & $10^{-4}$ & 50.2 & 78.4 & \highlight \textbf{56.5} & 18.9 & 42.2 & \highlight \textbf{22.3} \\
\cmidrule{2-9}
 & \multirow{3}{*}{4.00} & $\infty$ & 44.7 & 75.9 & 48.9 & 18.0 & 42.8 & 17.0 \\
 &  & $10^{-2}$ & 45.8 & 77.2 & 50.9 & 17.9 & 43.3 & 19.2 \\
 &  & $10^{-4}$ & 48.9 & 78.9 & 54.9 & 18.1 & 41.9 & 20.6 \\
\bottomrule
\end{tabular}
\end{table}

\begin{table}[t]
\centering
\footnotesize
\caption{Llama-3.2-3B - Fixed, Best checkpoint}
\begin{tabular}{lcccccccc}
\toprule
& & & \multicolumn{3}{c}{Math-500} & \multicolumn{3}{c}{Minerva} \\
Strategy & $\alpha$ & $\beta$ & Pass@1 & Pass@16 & Maj@16 & Pass@1 & Pass@16 & Maj@16 \\
\midrule
Base & -- & -- & -- & -- & -- & -- & -- & -- \\
\multirow{3}{*}{MH-LLM} & 1.0 & -- & 30.5 & 70.5 & 40.9 & 9.3 & 34.4 & 12.9 \\
 & 2.0 & -- & 44.4 & 74.1 & 50.1 & 18.2 & 41.3 & 21.1 \\
 & 4.0 & -- & 46.3 & 72.2 & 49.6 & \highlight \textbf{19.3} & 38.1 & 21.1 \\
Beam Search & -- & -- & 44.8 & -- & -- & 15.4 & -- & -- \\
\midrule
Task-Reward & -- & -- & \highlight \textbf{50.7} & 79.0 & 56.6 & 17.4 & 41.4 & 19.1 \\
Dist. Sharpen & -- & -- & 45.5 & 72.3 & 49.4 & 18.1 & 39.7 & 20.5 \\
\midrule
\multirow{12}{*}{Tilted Sampling} & \multirow{3}{*}{1.00} & $\infty$ & 38.5 & 78.0 & 48.6 & 11.4 & 38.0 & 14.0 \\
 &  & $10^{-2}$ & 46.6 & 79.8 & 55.2 & 16.2 & 43.5 & 18.1 \\
 &  & $10^{-4}$ & 49.9 & 78.1 & 55.3 & 18.7 & 42.4 & \highlight \textbf{21.9} \\
\cmidrule{2-9}
 & \multirow{3}{*}{1.25} & $\infty$ & 45.6 & 79.9 & 52.8 & 16.8 & 43.4 & 18.9 \\
 &  & $10^{-2}$ & 46.3 & \highlight \textbf{80.6} & 53.5 & 16.4 & 42.5 & 19.1 \\
 &  & $10^{-4}$ & 49.6 & 78.4 & 55.3 & 19.0 & \highlight \textbf{45.3} & 21.1 \\
\cmidrule{2-9}
 & \multirow{3}{*}{2.00} & $\infty$ & 46.4 & 77.8 & 51.7 & 18.5 & 42.6 & 20.5 \\
 &  & $10^{-2}$ & 45.7 & 78.3 & 52.8 & 17.2 & 42.3 & 20.2 \\
 &  & $10^{-4}$ & 50.0 & 79.9 & \highlight \textbf{56.8} & 18.5 & 40.9 & 20.7 \\
\cmidrule{2-9}
 & \multirow{3}{*}{4.00} & $\infty$ & 45.7 & 77.0 & 49.6 & 18.4 & 41.8 & 20.0 \\
 &  & $10^{-2}$ & 45.5 & 76.9 & 50.1 & 17.6 & 42.5 & 19.9 \\
 &  & $10^{-4}$ & 48.1 & 79.5 & 54.5 & 17.2 & 42.2 & 19.1 \\
\bottomrule
\end{tabular}
\end{table}

\begin{table}[t]
\centering
\scriptsize
\setlength{\tabcolsep}{3pt}
\caption{Qwen3-4B - Variable, Last checkpoint}
\begin{tabular}{lccccccccccc}
\toprule
& & & \multicolumn{3}{c}{AIME 2024} & \multicolumn{3}{c}{AIME 2025} & \multicolumn{3}{c}{HMMT 2025} \\
Strategy & $\alpha$ & $\beta$ & Pass@1 & Pass@16 & Maj@16 & Pass@1 & Pass@16 & Maj@16 & Pass@1 & Pass@16 & Maj@16 \\
\midrule
Base & -- & -- & 29.5 & 45.6 & 37.8 & 27.6 & 40.0 & 30.0 & 9.4 & 21.1 & 12.2 \\
\multirow{3}{*}{MH-LLM} & 1.0 & -- & {22.4} & {47.8} & {31.1} & {16.0} & {28.9} & {23.3} & {5.5} & {15.6} & {6.7} \\
 & 2.0 & -- & 28.1 & 55.6 & 31.1 & 23.7 & 33.3 & 25.6 & 7.5 & 21.1 & 10.0 \\
 & 4.0 & -- & 27.6 & 52.2 & 34.4 & 24.4 & 31.1 & 27.8 & 6.9 & 21.1 & 5.6 \\
Beam Search & -- & -- & 30.0 & -- & -- & 26.7 & -- & -- & 6.7 & -- & -- \\
\midrule
Task-Reward & -- & -- & 40.9 & 66.7 & 46.7 & 32.2 & 52.2 & 37.8 & 16.6 & 32.2 & 20.0 \\
Dist. Sharpen & -- & -- & \textcolor{red}{12.0} & \textcolor{red}{23.3} & \textcolor{red}{15.6} & \textcolor{red}{19.9} & \textcolor{red}{25.6} & \textcolor{red}{23.3} & \textcolor{red}{4.9} & \textcolor{red}{10.0} & \textcolor{red}{5.6} \\
\midrule
\multirow{12}{*}{Tilted Sampling} & \multirow{3}{*}{1.00} & $\infty$ & 29.5 & 45.6 & 37.8 & 27.6 & 40.0 & 30.0 & 9.4 & 21.1 & 12.2 \\
 &  & $10^{-2}$ & 36.9 & 56.7 & 45.6 & 29.3 & 48.9 & 33.3 & 12.1 & 26.7 & 16.7 \\
 &  & $10^{-4}$ & \highlight \textbf{45.3} & \highlight \textbf{71.1} & 45.6 & \highlight \textbf{35.8} & \highlight \textbf{56.7} & \highlight \textbf{42.2} & 16.0 & 32.2 & 15.6 \\
\cmidrule{2-12}
 & \multirow{3}{*}{1.25} & $\infty$ & 35.2 & 62.2 & 42.2 & 28.3 & 52.2 & 33.3 & 15.3 & 28.9 & \highlight \textbf{22.2} \\
 &  & $10^{-2}$ & 36.3 & 60.0 & 44.4 & 30.1 & 53.3 & 35.6 & 16.0 & 33.3 & 21.1 \\
 &  & $10^{-4}$ & 44.2 & 66.7 & 46.7 & 32.2 & 50.0 & 36.7 & \highlight \textbf{18.1} & 36.7 & 18.9 \\
\cmidrule{2-12}
 & \multirow{3}{*}{2.00} & $\infty$ & 22.2 & 41.1 & 25.6 & 18.6 & 32.2 & 22.2 & 8.6 & 17.8 & 8.9 \\
 &  & $10^{-2}$ & 31.4 & 54.4 & 35.6 & 25.6 & 48.9 & 26.7 & 14.2 & 31.1 & 14.4 \\
 &  & $10^{-4}$ & 44.4 & 64.4 & 47.8 & 34.0 & 55.6 & 36.7 & 15.5 & 32.2 & 13.3 \\
\cmidrule{2-12}
 & \multirow{3}{*}{4.00} & $\infty$ & 23.0 & 42.2 & 24.4 & 18.4 & 22.2 & 20.0 & 8.3 & 13.3 & 8.9 \\
 &  & $10^{-2}$ & 24.1 & 40.0 & 28.9 & 24.2 & 34.4 & 26.7 & 9.0 & 21.1 & 8.9 \\
 &  & $10^{-4}$ & 40.1 & 66.7 & \highlight \textbf{50.0} & 32.5 & \highlight \textbf{56.7} & 35.6 & 17.7 & \highlight \textbf{37.8} & 20.0 \\
\bottomrule
\end{tabular}
\end{table}

\begin{table}[t]
\centering
\scriptsize
\setlength{\tabcolsep}{3pt}
\caption{Qwen3-4B - Variable, Best checkpoint}
\begin{tabular}{lccccccccccc}
\toprule
& & & \multicolumn{3}{c}{AIME 2024} & \multicolumn{3}{c}{AIME 2025} & \multicolumn{3}{c}{HMMT 2025} \\
Strategy & $\alpha$ & $\beta$ & Pass@1 & Pass@16 & Maj@16 & Pass@1 & Pass@16 & Maj@16 & Pass@1 & Pass@16 & Maj@16 \\
\midrule
Base & -- & -- & 29.5 & 45.6 & 37.8 & 27.6 & 40.0 & 30.0 & 9.4 & 21.1 & 12.2 \\
\multirow{3}{*}{MH-LLM} & 1.0 & -- & {22.4} & {47.8} & {31.1} & {16.0} & {28.9} & {23.3} & {5.5} & {15.6} & {6.7} \\
 & 2.0 & -- & 28.1 & 55.6 & 31.1 & 23.7 & 33.3 & 25.6 & 7.5 & 21.1 & 10.0 \\
 & 4.0 & -- & 27.6 & 52.2 & 34.4 & 24.4 & 31.1 & 27.8 & 6.9 & 21.1 & 5.6 \\
Beam Search & -- & -- & 30.0 & -- & -- & 26.7 & -- & -- & 6.7 & -- & -- \\
\midrule
Task-Reward & -- & -- & 43.4 & 64.4 & 50.0 & 29.0 & 48.9 & 32.2 & 16.6 & 26.7 & 20.0 \\
Dist. Sharpen & -- & -- & 35.4 & 60.0 & 40.0 & 28.5 & 53.3 & 32.2 & 13.6 & 30.0 & 20.0 \\
\midrule
\multirow{12}{*}{Tilted Sampling} & \multirow{3}{*}{1.00} & $\infty$ & 29.5 & 45.6 & 37.8 & 27.6 & 40.0 & 30.0 & 9.4 & 21.1 & 12.2 \\
 &  & $10^{-2}$ & 36.4 & 58.9 & 44.4 & 29.5 & 48.9 & 33.3 & 12.6 & 22.2 & 18.9 \\
 &  & $10^{-4}$ & 45.3 & \highlight \textbf{71.1} & 45.6 & \highlight \textbf{35.8} & \highlight \textbf{56.7} & \highlight \textbf{42.2} & 16.0 & 32.2 & 15.6 \\
\cmidrule{2-12}
 & \multirow{3}{*}{1.25} & $\infty$ & 34.0 & 61.1 & 40.0 & 27.4 & 47.8 & 33.3 & 14.7 & 30.0 & 21.1 \\
 &  & $10^{-2}$ & 37.0 & 64.4 & 44.4 & 29.6 & 52.2 & 34.4 & 15.9 & 32.2 & \highlight \textbf{22.2} \\
 &  & $10^{-4}$ & \highlight \textbf{47.0} & 65.6 & \highlight \textbf{55.6} & 33.6 & 52.2 & 40.0 & 17.0 & 28.9 & 18.9 \\
\cmidrule{2-12}
 & \multirow{3}{*}{2.00} & $\infty$ & 34.6 & 62.2 & 43.3 & 28.1 & 45.6 & 35.6 & 15.3 & 32.2 & \highlight \textbf{22.2} \\
 &  & $10^{-2}$ & 36.7 & 70.0 & 43.3 & 26.9 & 53.3 & 28.9 & 14.0 & 33.3 & 20.0 \\
 &  & $10^{-4}$ & 45.0 & 67.8 & 47.8 & 33.0 & 52.2 & 37.8 & 16.1 & 32.2 & 16.7 \\
\cmidrule{2-12}
 & \multirow{3}{*}{4.00} & $\infty$ & 23.0 & 42.2 & 24.4 & 18.4 & 22.2 & 20.0 & 8.3 & 13.3 & 8.9 \\
 &  & $10^{-2}$ & 34.0 & 61.1 & 38.9 & 27.6 & 48.9 & 32.2 & 15.5 & \highlight \textbf{37.8} & 21.1 \\
 &  & $10^{-4}$ & 42.8 & 67.8 & 46.7 & 29.4 & 52.2 & 32.2 & \highlight \textbf{18.1} & 33.3 & 20.0 \\
\bottomrule
\end{tabular}
\end{table}

\end{document}